%% file: poseInduction.tex
\documentclass[10pt,twocolumn,letterpaper]{article}

\usepackage{iccv}
\usepackage{times}
\usepackage{epsfig}
\usepackage{graphicx}
\usepackage{amsmath}
\usepackage{amssymb}
\usepackage{bbm}
\usepackage{lib}
\usepackage{algorithm,algorithmic}
\usepackage{color}

\definecolor{minorEd}{rgb}{0,0,0}
\definecolor{majorEd}{rgb}{1,0,0}

\makeatletter
\newcommand{\HEADER}[1]{\ALC@it\underline{\textsc{#1}}\begin{ALC@g}}
\newcommand{\ENDHEADER}{\end{ALC@g}}
\makeatother

\newcommand\blfootnote[1]{%
  \begingroup
  \renewcommand\thefootnote{}\footnote{#1}%
  \addtocounter{footnote}{-1}%
  \endgroup
}

\usepackage[pagebackref=true,breaklinks=true,letterpaper=true,colorlinks,bookmarks=false]{hyperref}
\usepackage{booktabs}
\iccvfinalcopy 

\def\iccvPaperID{541} 

\begin{document}

\title{Pose Induction for Novel Object Categories}
\author{Shubham Tulsiani, Jo\~{a}o Carreira and Jitendra Malik\\
University of California, Berkeley\\
{\tt\small \{shubhtuls,carreira,malik\}@eecs.berkeley.edu}}

\maketitle

\begin{abstract}
We address the task of predicting pose for objects of unannotated object categories from a small seed set of annotated object classes. We present a generalized classifier that can reliably induce pose given a single instance of a novel category. In case of availability of a large collection of novel instances, our approach then jointly reasons over all instances to improve the initial estimates. We empirically validate the various components of our algorithm and quantitatively show that our method produces reliable pose estimates. We also show qualitative results on a diverse set of classes and further demonstrate the applicability of our system for learning shape models of novel object classes.
\end{abstract}

\blfootnote{Our implementations and trained models are available at \url{https://github.com/shubhtuls/poseInduction}}
\vspace{-1cm}
\input{introduction}
\input{induction}

\input{refinement}

\input{applications}

\input{conculsion}

\section*{Acknowledgements}
The authors would like to thank Jun-Yan Zhu for his valuable comments. This work was supported in part by NSF Award IIS-1212798 and ONR MURI-N00014-10-1-0933. Shubham Tulsiani was supported by the Berkeley fellowship and Jo\~{a}o Carreira was supported by the Portuguese Science Foundation, FCT, under grant SFRH/BPD/84194/2012. We gratefully acknowledge NVIDIA corporation for the donation of Tesla GPUs for this research.

{\small
\bibliographystyle{ieee}
\bibliography{poseInduction}
}

\end{document}

%% file: introduction.tex
\section{Introduction}

Class-based processing significantly simplifies tasks such as object segmentation \cite{hariharan2014simultaneous,carreira2012semantic}, reconstruction \cite{cashman2013dolphins,shapesKarTCM14,carvi14} and, more generally, the propagation of knowledge from class objects we have seen before to those we are seeing for the first time. Looking at the lion in \figref{fig1} humans can not only easily perceive its shape, but also tell that it is strong and dangerous, get an estimate of its weight and dimensions and even approximate age and gender. We get to know all of this because it is a lion like others we have seen before and that we know many facts about.

Despite its many virtues, class-based processing does not scale well. Learning predictors for all variables of interest -- figure-ground segmentation, pose, shape -- requires expensive manual annotations to be collected for at least dozens of examples per class and there are millions of classes. Consider again \figref{fig1} but now look at object A. The underlying structure in our visual world allows us to perceive a rich representation of this object despite encountering it for the first time. We can infer that it is probably hair that covers its surfaces -- we have seen plenty of hair-like materials before -- and that it has parts and determine their configuration by analogy with our own parts or with other animals. We are able to achieve this remarkable feat by leveraging commonalities across object categories via generalizable abstractions -- not only can we perceive that all the other animals in \figref{fig1} are ``right-facing", we can also transfer this notion to object A. This type of cross-category knowledge transfer has been successfully demonstrated before for properties such as materials \cite{varma2009statistical,CimpoiMV14}, parts \cite{torralba2007sharing,endres2012learning} and attributes \cite{lampert2009learning,farhadi2009describing}.

\begin{figure}[t!]
\includegraphics[width=0.45\textwidth]{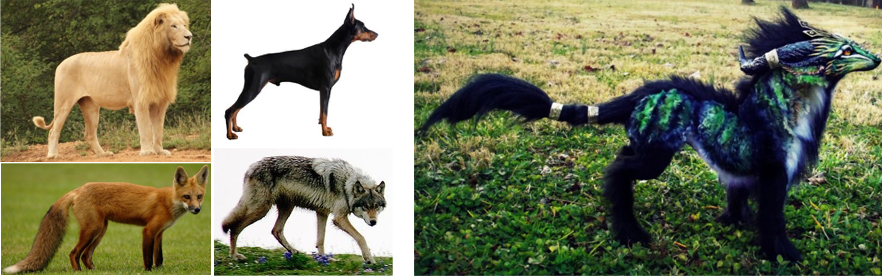}
\caption{Inductive pose inference for novel objects. Right : Novel object A. Left : instances from previously seen classes having similar pose as object A.}
\figlabel{fig1}
\end{figure}

\begin{figure*}[t!]
\includegraphics[width=\textwidth]{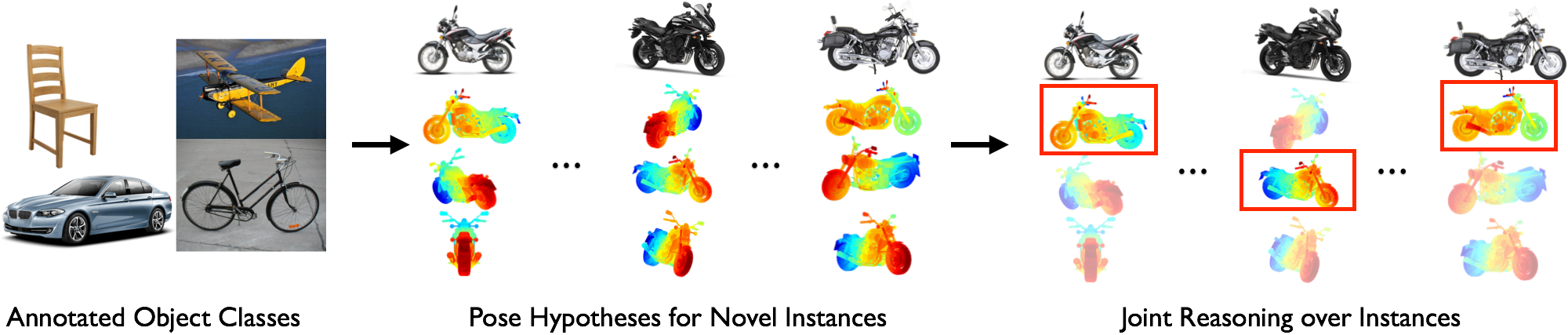}
\caption{Overview of our approach. We first induce pose hypotheses for novel object instances using a system trained over aligned annotated classes (\secref{poseInduction}). We then reason jointly over all instances of the novel object class to improve our pose predictions ( \secref{classInduction}).}
\figlabel{overview}
\end{figure*}

In this paper we define and attack the problem of predicting object poses across categories -- we call this pose induction. The first step of our approach, as highlighted in \figref{overview}, is to learn a generalizable pose prediction system from the given set of annotated object categories. Our main intuition is that most objects have appearance and shape traits that can be associated with a generalized notion of \textit{pose}. For example, the sentences ``I am in front of a car" or ``in front of a bus" or ``in front of a lion" are clear about where``I" am with respect to those objects. The reason for this may be that there is something generic in the way``frontality" manifests itself visually across different object classes -- e.g.``fronts" usually exhibit an axis of bilateral symmetry. Pushing this observation further leads to our solution: to align all the objects in a small \textit{seed} set of classes, by endowing them with set of 3D rotations in a consistent reference frame, then training pose predictors that generalize in a meaningful way to novel object classes.

This idea expands the current range of inferences that can be performed in a class-independent manner and allows us to reason about pose for every object without tediously collecting pose annotations. Such pose based reasoning can then inform a system about which directions objects are most likely to move in (usually ``front" or ``back") and hence allow it to get out of their way; it can help to identify how to place any object on top a surface in a stable way (by identifying the ``bottom" of the object). Ultimately, and the main motivation for this work, it provides important cues about the 3D shape of a novel object and may allow bypassing the existing need for ground truth keypoints in training data for state-of-the-art class-specific object reconstruction systems \cite{shapesKarTCM14,carvi14} -- we will present a proof of concept for this in Section \ref{sec:shapeModelling}.

\input{related}

%% file: related.tex
\vspace{3mm}
\noindent \textbf{Related Work.}
The problem of generalizing from a few examples \cite{tenenbaum2011grow} was already studied in ancient Greece and has become known as \textit{induction}. Early induction work in computer vision pursued feature sharing between different classes \cite{bart2005cross,torralba2007sharing}. One-Shot and Zero-Shot learning \cite{Li06OneShot, palatucci2009zero} also represent related areas of research where the task is to learn to predict labels from very few exemplars. Our work differs from these as, in constrast to these approaches, the \textit{few examples} we consider correspond to a small set of annotated object categories. In this sense, our approach is perhaps closer in style to \textit{attributes} \cite{farhadi2009describing,lampert2009learning}, which explicitly learn classifiers that are transversal to object classes and can hence be trained on a subset of object classes. Differently, our ``attributes" correspond to a dense discretization of the viewpoint manifold that implicitly aligns the shapes of all training object classes. Another relevant recent work, LSDA \cite{Hoffman14lsda} learns object detectors using a seed set of classes having bounding box annotations. Unlike our work, they leverage available data for a related task (classification) and frame the task as adapting classifiers to object detectors.

Pose estimation is crucial for developing a rich understanding of objects and is therefore an important component of systems for 3D reconstruction \cite{shapesKarTCM14,carreira2014virtual}, recognition \cite{osadchy2007synergistic,taigman2014deepface}, robotics \cite{simon1994real} and human computer interaction \cite{lepetit2004point,shotton2013real}. Traditional approaches to object pose estimation predicted instance pose in context of a corresponding shape model \cite{huttenlocher1990recognizing}. The task has recently evolved  to the prediction of category-level pose, a problem targeted by many recent methods \cite{vpsKpsTulsianiM14,pepik12dpm,ghodrati14viewpoint}. Motivated by Palmer's experiments which demonstrate common canonical frames for similar categories \cite{palmer1999vision}, we reason over cross-category pose - our work can be thought of as a natural extension in the current paradigm shift of pose prediction from instances/models to categories.

%% file: induction.tex
\section{Pose Induction for Object Instances}
\seclabel{poseInduction}

We noted earlier that humans have the ability to infer rich representations, including pose, even for previously unseen object classes. These observations demonstrate the applicability of human inductive learning as a mechanism to infer desired representations for new visual data. We explore the possibility of applications of such ideas to induce the notion of pose for \textcolor{minorEd}{previously} unseen object instances. More concretely, we assume pose annotations for some object classes and aim to infer pose for an object instance belonging to a \textcolor{minorEd}{different} object category. We describe our formulations and approach below.

\subsection{Formulation}
\seclabel{instanceFormulation}
Let $C$ denote the set of object categories with available pose annotations. We follow the pose estimation formulation of Tulsiani and Malik \cite{vpsKpsTulsianiM14} who characterize pose via $N_a=3$ euler angles  - azimuth ($\phi$), elevation($\varphi$) and cyclo-rotation($\psi$). We  \textcolor{minorEd}{discretize} the space of each angle in $N_\theta$ disjoint bins and frame the task of pose prediction as a classification problem to determine the angular bin for each euler angle. Let $\{x_i | i = 1 \ldots N_i\}$ denote the set of annotated instances, each with its object class $c_i \in C$, with pose annotations $(\phi_i,\varphi_i,\psi_i)$. The pose induction task is to predict the pose for a novel instance $x$ whose object class $c \notin C$.

\subsection{Approach}
\seclabel{instanceApproach}
We examine two different approaches for inducing pose for a novel instance - 1) the baseline approach of explicitly leveraging the inference mechanism for similar object classes and 2) our proposed approach of enforcing the inference mechanism to implicitly leverage similarities between object classes and thereby allowing generalization of inference to novel instances.

\vspace{2mm}
\noindent \textbf{Similar Classifier Transfer (SCT).}
We first describe the baseline approach which infers pose for instances of an unannotated class by explicitly using similarity to some annotated object category and obtaining predictions using a system trained for a visually similar class. To obtain a pose prediction system for the annotated classes $C$, we follow  the methodology of Tulsiani and Malik \cite{vpsKpsTulsianiM14} and train a VGG net \cite{Simonyan14c} based Convolutional Neural Network (CNN) \cite{neocognitron,LeCun1989} architecture with $|C|*N_a*N_\theta$ output units in the last layer. Each output unit corresponds to a particular object class, euler angle and angular bin - this CNN system shares most parameters across classes but has some class-specific parameters and disjoint output units.
Let $f(x;W_c)$ denote the pose prediction function for image $x$ and class-specific CNN weights $W_c$, then $f(x_i,W_{c_i})$ computes the probability distribution over angular bins for instance $i$ - the CNN is trained to minimize the softmax loss corresponding to the true pose label $(\phi_i,\varphi_i,\psi_i)$ and $f(x_i,W_{c_i})$. 

To predict pose for an instance $x$ with class $c \notin C$, this approach uses the prediction system for a visually similar class $c'$. We obtain the probability distribution over angular bins for this instance by computing $f(x,W_{c'})$. We then use the most likely hypothesis under this distribution as our pose estimate for the instance $x$.

\vspace{2mm}
\noindent \textbf{Generalized Classifier (GC).}
To infer properties for a novel instance, our proposed approach is to rely not only on the most similar visual object class, but also on general abstractions from all visual data - seeing a sheep for the first time, one would not just use knowledge of a specific class like cows, but also generic knowledge about four-legged animals. For example, the concept that pose of animals can be determined using generic part representations (head, torso etc.) can be learned if the annotations share a common canonical reference frame across classes and this notion can then be applied to novel related classes. These observations motivate us to consider an alternate approach, termed as Generalized Classifier (GC), where we train a system that exploits consistent visual similarities across object classes that \textcolor{minorEd}{coherently} change with the pose label. This approach not only bypasses the need for manually assigning a visually similar class, it can also potentially learn abstractions more generalizable to unseen data and therefore handle novel instances more robustly.

Concretely, we first obtain pose annotations across object classes wrt a common canonical frame (details described in experimental section) and train a category-agnostic pose estimation system. This implicitly enforces the CNN based pose estimation system to exploit similarities across object classes and learn common representations that may be useful to predict pose across object classes. We train a VGG net \cite{Simonyan14c} based CNN architecture with $N_a*N_\theta$ output units in the last layer - the units corresponds to a particular euler angle and angular bin are shared across all classes. Let $f(x;W)$ denote the pose prediction function for image $x$ and CNN weights $W$, then CNN is trained to minimize the softmax loss corresponding to the true pose label $(\phi_i,\varphi_i,\psi_i)$ and $f(x_i,W)$. To predict pose for an instance $x$ of an unannotated class $c$, we just compute $f(x;W)$ - the alignment of all annotated classes to a \textcolor{minorEd}{canonical} pose and implicit sharing of abstractions allow this system to generalize well to new object classes.

\subsection{Experiments}
\seclabel{instanceExperiments}
\noindent \textbf{Pose Annotations and Alignment.}
We evaluate the performance of our system on PASCAL VOC \cite{pascal-voc-2012} object categories. We obtain pose annotations for rigid categories via the PASCAL3D+ \cite{pascal3d} dataset which annotates instances in PASCAL VOC and Imagenet dataset with their euler angles. The notion of a global viewpoint is challenging to define for various animal categories in PASCAL VOC and we apply SfM-based techniques on ground truth keypoints to obtain the torso pose. We use keypoints annotations provided by Bourdev \etal \cite{bourdevECCV10} followed by rigid factorization \cite{carvi14} to obtain viewpoint for non-rigid pascal classes. The PASCAL3D+ annotations assume a canonical reference frame across classes - objects are laterally symmetric across X axis and face frontally in the canonical pose. We obtain similarly aligned reference frames for other object classes by aligning the SfM models to adhere to this constraint.

\vspace{2mm}
\noindent \textbf{Evaluation Setup.}
We held out pose annotations for four object classes - bus, dog, motorbike and sheep. We then finetuned the CNN systems, after initializing weights using a pretrained model for Imagenet \cite{ImageNet} classification, corresponding to the two approaches described above using pose annotations for the remaining 16 classes obtained via PASCAL3D+ or PASCAL VOC keypoint labels.

To evaluate the performance of our system for rigid objects, we used the \textbf{$Acc_{\theta}$} metric \cite{vpsKpsTulsianiM14} which measures the fraction of instances whose predicted viewpoint is within a fixed threshold of the correct viewpoint (we use $\theta = \frac{\pi}{6}$). The `ground-truth' viewpoint obtained for some classes via SfM techniques is often noisy and the above metric which works well for exact annotations needs to be altered. To evaluate the system's performance for these classes, we use an auxiliary task of predicting the  `frontal/left/right/rear-facing' label available in PASCAL VOC for these objects. We use our predicted azimuth for these objects and infer the `frontal/left/right/rear-facing'  label based on the predicted azimuth. We denote the metric that measures accuracy at this auxiliary task as \textbf{$Acc_{v}$}.

\vspace{2mm}
\noindent \textbf{Results.}
We report the performance our baseline and proposed approach in Table \ref{table:instanceEval}. For the SCT method, we used the weights from car, bicycle, cat and cow prediction systems to predict pose for bus, motorbike, dog and sheep respectively since these correspond to the visually most similar classes with available annotations. We note that the predictions using both approaches are often very close to the actual object pose and are \textcolor{minorEd}{significantly} better than chance. We also observe that training a generalized prediction system is better than explicitly using a similar class (except for motorbike, where the bicycle class is very similar). This is perhaps because sharing of parameters and output units across classes enables learning shared abstractions that generalize better to novel classes.

\begin{table}[htb!]
\centering

\begin{tabular}{lcc|cc}
\toprule
 & \multicolumn{2}{c}{$Acc_{\frac{\pi}{6}}$} & \multicolumn{2}{c}{$Acc_{v}$}\tabularnewline
\midrule
Approach & bus & mbike & dog & sheep\tabularnewline
\hline 
SCT & {\small{}0.50} & \textbf{\small{}0.58} & \textbf{\small{}0.75} & {\small{}0.58}\tabularnewline
GC & \textbf{\small{}0.80} & {\small{}0.55} & {\small{}0.74} & \textbf{\small{}0.78}\tabularnewline
\bottomrule
\end{tabular}

\vspace{2mm}
\caption{Performance of our approaches for various novel object classes.}
\label{table:instanceEval}
\end{table}

We have \textcolor{minorEd}{described} a methodology that aims to provide a richer description, in particular pose, given a single instance belonging to an novel class. We note that though human levels of precision and understanding for novel objects are still far away, the results imply that we can reliably predict pose without requiring training annotations, which is a step in the direction of visual systems capable of dealing with new instances.

\vspace{2mm}
\noindent \textbf{Importance of Similar Object Categories.} To further gain insight into our prediction system, we focused on the `bus' object category and trained two additional networks for the GC method by holding out `car' and `chair' respectively (in addition to the four held out categories above). In comparison to $Acc_{\frac{\pi}{6}}$ = 0.80, the $Acc_{\frac{\pi}{6}}$ measure for bus in these two cases was 0.73 and 0.81 respectively. The observed drop by holding out `car' confirms our intuition regarding the importance of similar object categories in the seed set.

%% file: refinement.tex
\section{Pose Induction for Object Categories}
\seclabel{classInduction}

When reasoning over a single instance of a novel category, any system, including the approaches in \secref{poseInduction}, can only rely on inference and abstractions on previously seen visual data. However, if given at once a collection of instances belonging to the new category,  we can infer pose for all instances of the object class under consideration while reasoning jointly over all of their poses. This allows us to go beyond isolated reasoning for each instance and leverage the collection of images to jointly reason over and infer pose for all instances of the object class under consideration. Tackling the problem of inducing pose at a category level is particularly relevant as pose annotations for objects are far more tedious to collect than class labels -- there are significantly more datasets with annotated classes than pose. Our method allows us to augment these available datasets with a notion of pose for each object. Our method can also be used in a completely unsupervised setting to infer pose for consistent visual clusters over instances that visual knowledge extraction systems like NEIL \cite{chen_iccv13} automatically discover.

One possible approach to reasoning jointly is to explicitly infer intra-class correspondences, predict relative transformations and augment these with the induced instance predictions to obtain more informed pose estimates for each instance. However, the task of discovering correspondences across  instances that differ in both pose and appearance, is a particularly challenging one and has been demonstrated only in limited pose and appearance variability \cite{flowweb,Singh2012DiscPat}. Our proposed approach provides a simpler but more robust way of leveraging the image collection. We build on the intuition that instances with similar spatial distributions of parts are close on the pose manifold. We define a similarity measure that captures this intuition and encourage similar instances to have similar pose predictions.

\input{vpPredFig}

\begin{algorithm}
\caption{Joint Pose Induction}
    \begin{algorithmic}
    \HEADER{Initialization}
      \FOR{i in test instances}
          \STATE Predict pose distribution $F(x_i;W)$ (\secref{poseInduction})
          \STATE Compute K pose hypotheses and likelihood scores $\{ (R_{ik},\beta_{ik})| k \in \{ 1,..,K \} \}$ using $F(x_i;W)$
          \STATE Compute similar instances $N_i$ using $F_i$ (eq \ref{eq:optFeature})
          \STATE $z_i  \gets \underset{k}{\text{argmax}} \beta_{ik}$
        \ENDFOR
    \ENDHEADER

    \STATE
    
    \HEADER{Pose Refinement}
      \STATE $\forall i$, Update $z_i$ (eq \ref{eq:icmUpdate}) until  convergence
    \ENDHEADER

  \end{algorithmic}
\label{alg:induction}
\end{algorithm}

\input{inductionPredsAll}
\input{inductionPredsSupp2}

\subsection{Approach}
We first obtain multiple pose hypotheses for each instance by obtaining a diverse set of modes from the distribution predicted by the system described in \secref{poseInduction} . We then frame the joint pose prediction task as that of selecting a hypothesis for each instance while taking into consideration the prediction confidence score as well as pose consistency with similar instances. We describe our formulation in detail below.

\vspace{2mm}
\noindent \textbf{Instance Similarity.}
For each instance $i$, we obtain a set of instances $N_i$ whose feature representations are similar to instance $i$. Our feature representation for an instance is motivated by the observation that each channel in a higher-layer of a CNN can be reasoned as encoding a spatial likelihood of abstract parts. Let $C_i(x,y,k)$ denote the instance's convolutional feature response for channel $k$ at location $(x,y)$, our feature representation $F_i$ is as follows.

\begin{gather}
\label{eq:optFeature}
F_i(\cdot,\cdot,k) = \frac{\sigma(C_i(\cdot,\cdot,k))}{\| \sigma(C_i(\cdot,\cdot,k)) \|_1}
\end{gather}

The above, where $ \sigma(\cdot)$ represents a sigmoid function, encodes each instance via the normalized spatial likelihood of these `parts'. We use histogram intersection over these representations as a similarity measure between two instances and obtain the set of neighbors $N_i$ for each instance.

\vspace{2mm}
\noindent \textbf{Unaries.}
For each instance $i$, we obtain $K$ distinct pose hypotheses $\{ R_{ik} | k \in \{ 1,..,K \} \}$ along with the corresponding log-likelihood scores  $ \beta_{ik} $. By $Z_i \in \{ 1,..,K \}$, we denote the random variable which corresponds to the pose hypothesis we select for instance $i$. The log-likelihood scores for each pose hypothesis act as the unary likelihood terms.
\begin{gather}
\label{eq:optUnary}
P_u(Z_i = z_i) \propto e^{\beta_{iz_i}}
\end{gather}

\vspace{2mm}
\noindent \textbf{Pose Consistency.}
Let $\Delta(R_1,R_2) = \frac{ \| log(R_1^TR_2)\|_F}{\sqrt{2}}$ denote the geodesic distance between rotation matrices $R_1,R_2$ and $\mathcal{I}$ denote the indicator function. We model the consistency likelihood term as the fraction of instances in $N_i$ with a similar pose.
\begin{gather}
\label{eq:optConsistency}
P_c(Z_i = z_i) \propto \frac{\underset{j \in N_i}\sum \mathcal{I}(\Delta(R_{iz_i},R_{jz_j}) < \delta)}{|N_i|}
\end{gather}
While this formulation encourages similar pose estimates for neighbors, it is biased towards more 'popular' pose estimates (if the dataset has more front facing bikes, it is more likely to find neighbors for the corresponding pose hypothesis). Motivated by the recent work of Isola \etal \cite{crispBoundaries}, who use Pointwise Mutual Information \cite{fano1961} (PMI) to counter similar biases, we normalize by the likelihood of randomly finding similar pose estimates for neighbors to yield -

\begin{gather}
\label{eq:optConsistencyPmi}
P_c(Z_i = z_i) \propto \frac{\underset{j \in N_i}\sum \mathcal{I}(\Delta(R_{iz_i},R_{jz_j}) < \delta)}{\underset{j}\sum \mathcal{I}(\Delta(R_{iz_i},R_{jz_j}) < \delta)}
\end{gather}

\vspace{2mm}
\noindent \textbf{Formulation.} \textcolor{minorEd}{$P_u$} favors the pose hypotheses that are scored higher by the instance pose \textcolor{minorEd}{induction} system and $P_c$, weighted by a factor of $\lambda$, leads to a higher joint probability if predicted pose in consistent with pose for similar instances. We finally combine these two likelihood terms to model the likelihood for the pose hypotheses for a given instance.
\begin{gather}
\label{eq:optTotal}
P(Z_i = z_i) \propto P_u(z_i) P_c(z_i)^\lambda
\end{gather}

\vspace{2mm}
\noindent \textbf{Inference.}
We aim to infer the MAP estimates $z_i^*$ for all instances to give us a pose prediction via joint reasoning over all instances. We use iterative updates and at each step, we condition on all the unknown variables except a particular $Z_i$; the update for assignment $z_i$ as follows -

\begin{multline}
\label{eq:icmUpdate}
z_i  = \underset{k}{\text{argmax}} (\beta_{ik}  + \lambda log(\underset{j \in N(i)}\sum \mathcal{I}(\Delta(R_{ik},R_{jz_j}) < \delta)) - \\ \lambda log(\underset{j}\sum \mathcal{I}(\Delta(R_{ik},R_{jz_j}) < \delta)))
\end{multline}

Our overall method, as summarized in Algorithm \ref{alg:induction}, computes pose estimates for every instance of a novel object class given a large collection of instances.

\subsection{Experiments}
The aim of the experiments is twofold - 1) to demonstrate the benefits of jointly reasoning over all instances of the class and 2) to show that a spatial feature representation capturing abstract parts, as defined in eq \ref{eq:optFeature}, yields better performance than alternatives for improving pose estimates. We follow the experimental setup previously described in \secref{instanceExperiments} and build on the `GC' approach. Our method using spatial features (from Conv5 of VGG net) is denoted as `GC+c5' and the  alternate similarity representation using fc7 features from VGG net is denoted as `GC+fc7'. We visualize the performance of our system in \figref{viewpointPreds} where the columns show $15^{th} - 75^{th}$ percentile instances, when sorted in terms of error. We observe that the predictions are accurate even around the $60^{th} - 75^{th}$ percentile regime.

\begin{table}[t!]
\centering

\begin{tabular}{lcc|cc}
\toprule
 & \multicolumn{2}{c}{$Acc_{\frac{\pi}{6}}$} & \multicolumn{2}{c}{$Acc_{v}$}\tabularnewline
\midrule
Approach & bus & mbike & dog & sheep\tabularnewline
\midrule
GC & {\small{}0.80} & {\small{}0.55} & \textbf{\small{}0.74} & {\small{}0.77}\tabularnewline
GC+fc7 & {\small{}0.76} & {\small{}0.51} & {\small{}0.73} & {\small{}0.75}\tabularnewline
GC+c5 & \textbf{\small{}0.86} & \textbf{\small{}0.60} & \textbf{\small{}0.74} & \textbf{\small{}0.79}\tabularnewline
\toprule
\end{tabular}

\vspace{2mm}
\caption{Joint \textcolor{minorEd}{reasoning} for Pose Induction.}
\label{table:classEval}
\end{table}

\begin{table}[th!]
\centering

\begin{tabular}{lcc|cc}
\toprule
 & \multicolumn{2}{c}{$Acc_{\frac{\pi}{6}}$} & \multicolumn{2}{c}{$Acc_{v}$}\tabularnewline
\midrule
\textbf{Setting} & bus & mbike & dog & sheep\tabularnewline
\midrule
All & {\small{}0.86} & {\small{}0.60} & {\small{}0.74} & {\small{}0.79}\tabularnewline
Confident & {\small{}0.97} & {\small{}0.76} & {\small{} 0.89} & {\small{} 0.90}\tabularnewline
\toprule
\end{tabular}

\vspace{2mm}
\caption{Performance for Confident Predictions.}
\label{table:classConfidence}
\end{table}

We see that the results in Table \ref{table:classEval} clearly support our two main hypotheses - that given multiple instances of a novel category, jointly reasoning over all of them improves the induced pose estimates and that the feature representation we described further improves performance.

An additional result that we show in Table \ref{table:classConfidence} is that if we rank the predictions by confidence (eq. \ref{eq:optTotal}) and take the top third confident predictions, error rates are significantly reduced. This means that the pose induction system has the desirable property of having low confidence when it fails. As we demonstrate later, for various applications \eg shape model learning, we might only need accurate pose estimates for a subset of instances and this result allows us to automatically find that subset  by selecting the top few in terms of confidence.

\subsection{Qualitative Results}
The evaluation setup so far has focused on PASCAL VOC object classes because of readily available annotations to measure performance. However, the aim of our method is to be able to infer pose annotations for any novel object category. We can qualitatively demonstrate the applicability of our approach for diverse classes using the Imagenet object classes.  \figref{inductionPreds} shows the predictions of our method for several classes for which we do not use any pose annotations (we use randomly selected instances from the top third, in terms of \textcolor{minorEd}{prediction} confidence, to visualize the predictions in  \figref{inductionPreds}). It is clear that the system performs well on animals in general as well as for other classes related to the initial training set (eg. golfcart, motorbike). While we can often infer a meaningful \textcolor{minorEd}{representation} of pose even for some classes rather different from the initial training classes e.g. hammer,  object categories which differ drastically from the annotated seed set (eg. jellyfish, vacuum cleaner) are the principal failure modes as illustrated in \figref{inductionPredsSuppFailure}.

%% file: vpPredFig.tex
\newcommand{\gtViewWidthVp}{0.19}
\newcommand{\gtViewFormatVp}{png}
\begin{figure*}[t!]
\includegraphics[width=\gtViewWidthVp\textwidth]{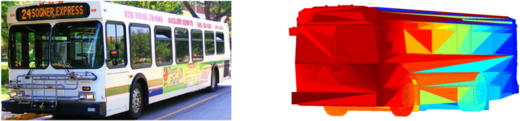} \hfill
\includegraphics[width=\gtViewWidthVp\textwidth]{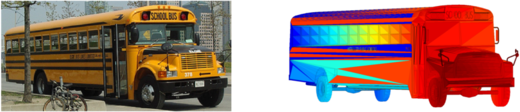} \hfill
\includegraphics[width=\gtViewWidthVp\textwidth]{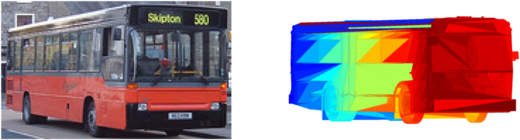}
\includegraphics[width=\gtViewWidthVp\textwidth]{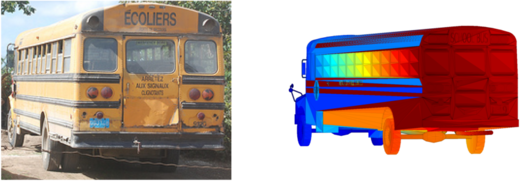} \hfill
\includegraphics[width=\gtViewWidthVp\textwidth]{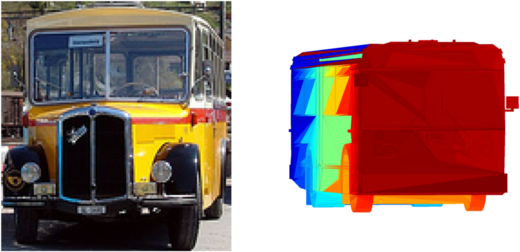} \hfill

\includegraphics[width=\gtViewWidthVp\textwidth]{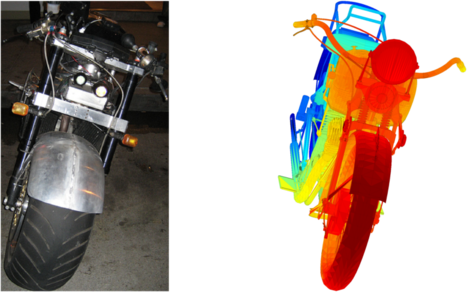} \hfill
\includegraphics[width=\gtViewWidthVp\textwidth]{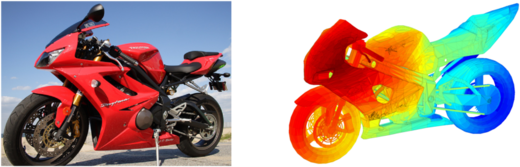} \hfill
\includegraphics[width=\gtViewWidthVp\textwidth]{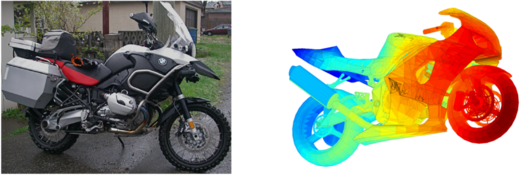}
\includegraphics[width=\gtViewWidthVp\textwidth]{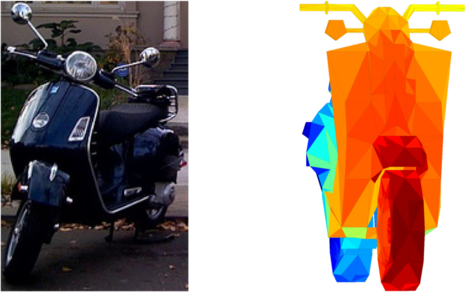} \hfill
\includegraphics[width=\gtViewWidthVp\textwidth]{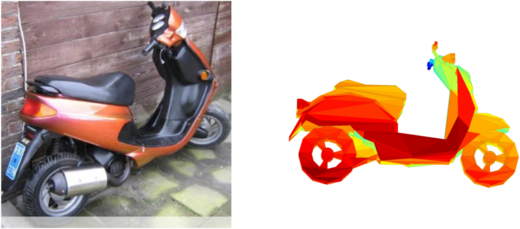} \hfill

\caption{Viewpoint predictions for unoccluded groundtruth instances using our full system ('GC+sim').  The columns show 15th, 30th, 45th, 60th and 75th percentile instances respectively in terms of the error. We visualize the predictions by rendering a 3D model using our predicted viewpoint.}
\figlabel{viewpointPreds}
\end{figure*}

%% file: inductionPredsAll.tex
\newcommand{\gtViewWidth}{0.186}
\newcommand{\gtViewFormat}{png}
\begin{figure*}[th!]
\centering
\setlength{\tabcolsep}{1.8pt}
\renewcommand{\arraystretch}{2.0}
\begin{tabular}{ccccc}
\includegraphics[width=\gtViewWidth\textwidth]{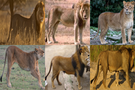} &
\includegraphics[width=\gtViewWidth\textwidth]{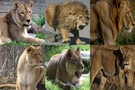} &
\includegraphics[width=\gtViewWidth\textwidth]{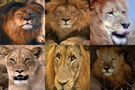} &
\includegraphics[width=\gtViewWidth\textwidth]{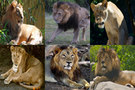} &
\includegraphics[width=\gtViewWidth\textwidth]{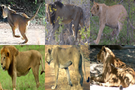} \\

\includegraphics[width=\gtViewWidth\textwidth]{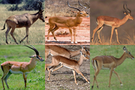} &
\includegraphics[width=\gtViewWidth\textwidth]{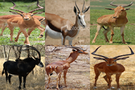} &
\includegraphics[width=\gtViewWidth\textwidth]{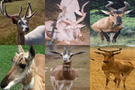} &
\includegraphics[width=\gtViewWidth\textwidth]{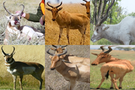} &
\includegraphics[width=\gtViewWidth\textwidth]{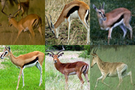} \\

\includegraphics[width=\gtViewWidth\textwidth]{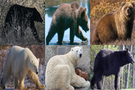} &
\includegraphics[width=\gtViewWidth\textwidth]{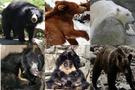} &
\includegraphics[width=\gtViewWidth\textwidth]{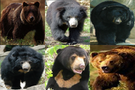} &
\includegraphics[width=\gtViewWidth\textwidth]{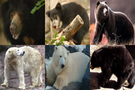} &
\includegraphics[width=\gtViewWidth\textwidth]{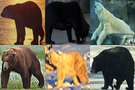} \\

\includegraphics[width=\gtViewWidth\textwidth]{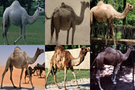} &
\includegraphics[width=\gtViewWidth\textwidth]{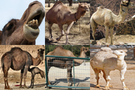} &
\includegraphics[width=\gtViewWidth\textwidth]{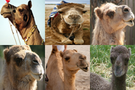} &
\includegraphics[width=\gtViewWidth\textwidth]{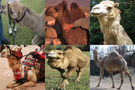} &
\includegraphics[width=\gtViewWidth\textwidth]{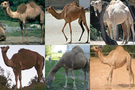} \\

\includegraphics[width=\gtViewWidth\textwidth]{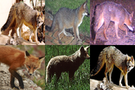} &
\includegraphics[width=\gtViewWidth\textwidth]{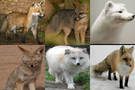} &
\includegraphics[width=\gtViewWidth\textwidth]{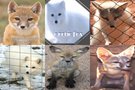} &
\includegraphics[width=\gtViewWidth\textwidth]{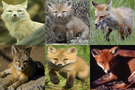} &
\includegraphics[width=\gtViewWidth\textwidth]{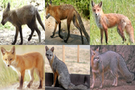} \\

\includegraphics[width=\gtViewWidth\textwidth]{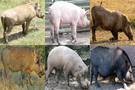} &
\includegraphics[width=\gtViewWidth\textwidth]{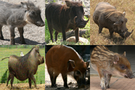} &
\includegraphics[width=\gtViewWidth\textwidth]{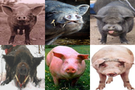} &
\includegraphics[width=\gtViewWidth\textwidth]{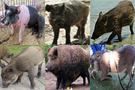} &
\includegraphics[width=\gtViewWidth\textwidth]{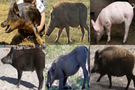} \\

\includegraphics[width=\gtViewWidth\textwidth]{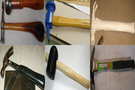} &
\includegraphics[width=\gtViewWidth\textwidth]{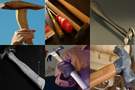} &
\includegraphics[width=\gtViewWidth\textwidth]{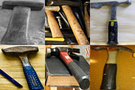} &
\includegraphics[width=\gtViewWidth\textwidth]{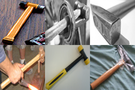} &
\includegraphics[width=\gtViewWidth\textwidth]{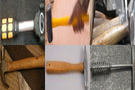} \\

\includegraphics[width=\gtViewWidth\textwidth]{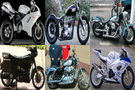} &
\includegraphics[width=\gtViewWidth\textwidth]{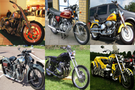} &
\includegraphics[width=\gtViewWidth\textwidth]{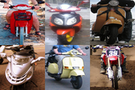} &
\includegraphics[width=\gtViewWidth\textwidth]{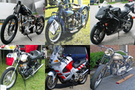} &
\includegraphics[width=\gtViewWidth\textwidth]{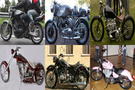} \\

\includegraphics[width=\gtViewWidth\textwidth]{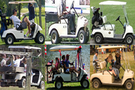} &
\includegraphics[width=\gtViewWidth\textwidth]{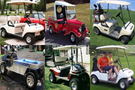} &
\includegraphics[width=\gtViewWidth\textwidth]{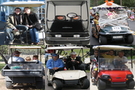} &
\includegraphics[width=\gtViewWidth\textwidth]{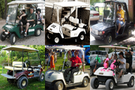} &
\includegraphics[width=\gtViewWidth\textwidth]{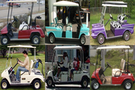}

\end{tabular}
\caption{Viewpoint predictions for novel object classes without any pose annotations. The columns show randomly selected instances whose azimuth is predicted to be around $\frac{-\pi}{2} (\text{right-facing}), \frac{-\pi}{4}, 0(\text{front-facing}), \frac{\pi}{4}, \frac{\pi}{2}(\text{left-facing})$ respectively.}
\figlabel{inductionPreds}
\end{figure*}

%% file: inductionPredsSupp2.tex
\newcommand{\figTwoClassOne}{jellyfish}
\newcommand{\figTwoClassTwo}{vacuum}
\begin{figure*}[th!]
\centering
\setlength{\tabcolsep}{1.8pt}
\renewcommand{\arraystretch}{2.0}
\begin{tabular}{ccccc}
\includegraphics[width=\gtViewWidth\textwidth]{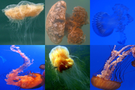} &
\includegraphics[width=\gtViewWidth\textwidth]{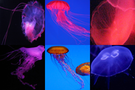} &
\includegraphics[width=\gtViewWidth\textwidth]{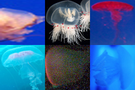} &
\includegraphics[width=\gtViewWidth\textwidth]{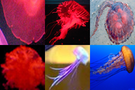} &
\includegraphics[width=\gtViewWidth\textwidth]{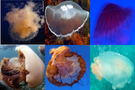} \\

\includegraphics[width=\gtViewWidth\textwidth]{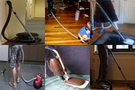} &
\includegraphics[width=\gtViewWidth\textwidth]{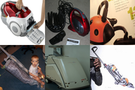} &
\includegraphics[width=\gtViewWidth\textwidth]{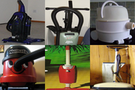} &
\includegraphics[width=\gtViewWidth\textwidth]{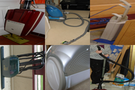} &
\includegraphics[width=\gtViewWidth\textwidth]{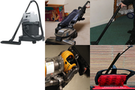}

\end{tabular}
\caption{Failure modes. Our method is unable to induce pose for object classes which drastically differ from the annotated seed set. The columns show randomly selected instances whose azimuth is predicted to be around $\frac{-\pi}{2} (\text{right-facing}), \frac{-\pi}{4}, 0(\text{front-facing}), \frac{\pi}{4}, \frac{\pi}{2}(\text{left-facing})$ respectively.}
\figlabel{inductionPredsSuppFailure}
\end{figure*}

%% file: applications.tex
\section{Shape Modelling for Novel Object Classes}
\label{sec:shapeModelling}

Acquiring shape models for generic object categories is an integral component of perceiving scenes with a rich 3D representation. The conventional approach to acquiring shape models includes leveraging human experts to build 3D CAD models of various shapes. This approach, however, cannot scale to a large number of classes while capturing the wildly different shapes in each object class. Learning based approaches which also allow shape deformations \cite{blanz1999morphable} provide an alternative solution but typically rely on some 3D initialization \cite{cashman2013dolphins}. Kar \etal \cite{shapesKarTCM14} recently showed that these models can be learned using annotations for only object silhouettes and a set of keypoints. These requirements, while an improvement over previous approaches, are still prohibitive for deploying similar approaches on a large scale. Enabling such approaches to learn shape models in the wild - given nothing but a set of instances, is an important \textcolor{minorEd}{endeavor} as it would allow us to scale shape model \textcolor{minorEd}{acquisition} to a large set of objects.

We take a step towards this goal using our pose induction system - we demonstrate that it is possible to learn shape models for a novel object category using just object silhouette annotations. We build on the formulation by Kar \etal \cite{shapesKarTCM14} and note that they mainly used keypoint annotations to estimate camera projection parameters and that these can be initialized using our induced pose as well. We briefly review their formulation and describe our modifications that allow us to learn shape models without keypoint annotations.

\begin{figure}[t!]
\includegraphics[width=0.9\linewidth]{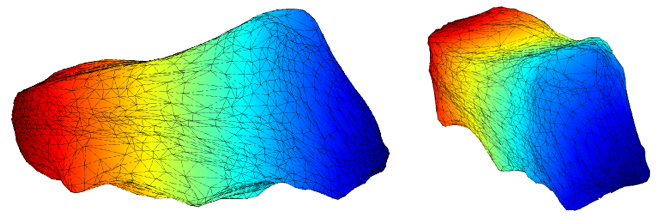} \\
\includegraphics[width=0.9\linewidth]{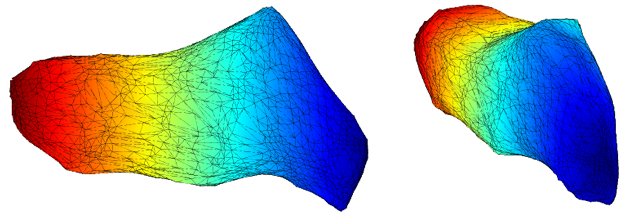} \\
\includegraphics[width=0.9\linewidth]{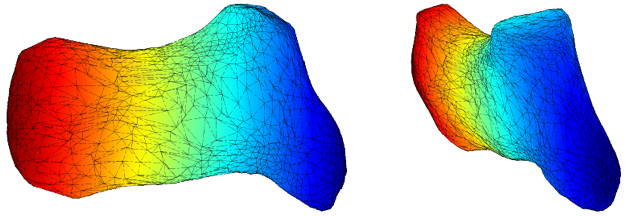} \\

\caption{Mean shape models learnt for motorbike using a) top : all pose induction estimates b) mid : most confident pose induction estimates c) bottom : ground-truth keypoint annotations.}
\figlabel{recons}
\end{figure}

\vspace{2mm}
\noindent \textbf{Formulation.}
Let $P_i = (R_i,c_i,t_i)$ represent the projection parameters (rotation, scale and translation) for the $i^{th}$ instance. Kar \etal obtain these using the annotated keypoints and we instead initialize the scale, translation parameters using bounding box scale, location and the rotation using our induced pose. 
Their shape model $M = (\overline{S},V)$ consists of a mean shape $\overline{S}$ and linear deformation bases $V = \{ V_1,.,V_K \}$. The energies used in their formulation enforce that the shape for an instance is consistent with its silhouette $(E_s, E_c)$, shapes are locally consistent $(E_l)$, normals vary smoothly $(E_n)$ and the deformation parameters are small $(\|\alpha_{ik}V_k\|_F^{2})$ (they also use a keypoint based energy $E_{kp}$ which we ignore). We refer the reader to \cite{shapesKarTCM14} for details regarding the optimization and formulations of shape energies. While Kar \etal only optimize over shape model and deformation parameters, we note that since our projection parameters are noisy, we should also refine them to minimize the energy. Therefore, we minimize the objective mentioned in eq. \ref{eq:optimization} over the shape model, deformation parameters as well as projection parameters (initialized using the induced pose) to learn shape models of a novel object class using just silhouette annotations.

\begin{equation}
\begin{aligned}
\label{eq:optimization}
& \underset{\bar{S},V,\alpha, P}{\text{min}}
E_{l}(\bar{S},V)+\underset{i}{\sum}(E_{s}^i+E_{c}^i+E_{n}^i+ \underset{k}{\sum}(\|\alpha_{ik}V_k\|_F^{2}))\\
& \text{subject to:} ~
S^i = \bar{S} + \underset{k}{\sum}\alpha_{ik} V_k
\end{aligned}
\end{equation}

\noindent \textbf{Results.}
We use the unoccluded instances of the class motorbike to demonstrate the applicability of our pose induction system for shape learning. Since we are interested in learning a shape model for the class, we can ignore some object instances for which we are uncertain regarding pose. As shown in table \ref{table:classConfidence}, we can use the subset of most confident pose estimates to get a higher level of precision. \figref{recons} shows that our model learnt without any keypoint annotation is quite similar to the model learnt by Kar \etal using full annotations and that using the subset of instances with confident pose induction predictions substantially improves shape models. The learnt model demonstrates that our pose induction system makes it is feasible to learn shape models for novel object classes without requiring keypoint annotations. This not only qualitatively verifies the reliability of our pose induction estimates, it also signifies an important step towards automatically learning shape representations from images.

%% file: conculsion.tex
\section{Conclusion}

We have presented a system which leverages available pose annotations for a small set of seed classes and can induce pose for a novel object class. We have empirically shown that the system performs well given a single instance of a novel class and that this performance is significantly improved if we reason jointly over multiple instances of that class, when available. We have also shown that our pose induction system enables learning shape representations for object classes without any keypoint/3D annotations required by previous methods. Our qualitative results on Imagenet further demonstrate that this approach generalizes to a large and diverse set of object classes. 
